\documentclass[letterpaper, 10 pt, journal, twoside]{IEEEtran}

\ifCLASSINFOpdf
\else
\fi

\usepackage{times}
\usepackage{epsfig}
\usepackage{graphicx}
\usepackage{amsmath}
\usepackage{amssymb}
\usepackage{balance}
\usepackage[percent]{overpic}
\usepackage{pifont}%
\usepackage{wasysym}
\usepackage[dvipsnames]{xcolor}
\usepackage[utf8]{inputenc} %
\usepackage[T1]{fontenc}    %
\definecolor{mypink1}{rgb}{0.78, 0.21, 0.13}
\usepackage[colorlinks=true]{hyperref}       %
\hypersetup{
    colorlinks=true,
    linkcolor=mypink1,
    filecolor=magenta,  
    citecolor=black
    }

\usepackage{url}            %
\usepackage{amsfonts}       %
\usepackage{nicefrac}       %
\usepackage{microtype}      %
\usepackage{tabularx}
\usepackage{booktabs}
\usepackage{graphicx}
\usepackage[export]{adjustbox}
\usepackage{xcolor}
\usepackage{overpic}
\usepackage{multirow}
\usepackage{times}
\usepackage{epsfig}
\usepackage{graphicx}
\usepackage{amsmath}
\usepackage{amssymb}
\usepackage{balance}
\usepackage[percent]{overpic}
\usepackage{pifont}
\usepackage{wasysym}
\usepackage{booktabs}

\usepackage{booktabs}
\usepackage{balance}
\usepackage{multirow}
\usepackage{overpic}
\usepackage{cite}
\usepackage[font=small,labelfont=bf]{caption}
\usepackage{color}
\usepackage{pifont}
\makeatletter
\@namedef{ver@everyshi.sty}{}
\makeatother
\usepackage{tikz}
\usetikzlibrary{patterns}

\usepackage{colortbl}
\usepackage{pgfplots}
\pgfplotsset{compat=1.17}
\usepackage{tabularx}
\usepackage{arydshln}
\usepackage{amssymb}%
\usepackage{pifont}%

\usepackage{wasysym}%
\let\labelindent\relax
\usepackage{enumitem}
\usepackage{wrapfig}
\usepackage{xspace}
\usepackage{tabu}
\usepackage[super]{nth}
\usepackage{scalefnt}
\usepackage{dsfont}
\usepackage{graphicx,calc}
\usepackage{subcaption}
\usepackage{algorithm}
\usepackage{algpseudocode}

\definecolor{codegreen}{rgb}{0,0.6,0}
\definecolor{codegray}{rgb}{0.5,0.5,0.5}
\definecolor{codepurple}{rgb}{0.58,0,0.82}
\definecolor{backcolour}{rgb}{0.95,0.95,0.92}

\definecolor{ourPurple}{HTML}{9673A6}
\definecolor{ourOrange}{HTML}{D79B00}
\definecolor{ourGreen}{HTML}{82B366}
\definecolor{ourRed}{HTML}{B85450}
\definecolor{personColor}{HTML}{0000FF}
\definecolor{bgColor}{HTML}{bed4f3}

\definecolor{darkgreen}{RGB}{0,153,51}
\definecolor{linkgreen}{RGB}{52,130,48}
\definecolor{LightCyan}{rgb}{0.87,0.92,0.96}
\definecolor{m_green}{RGB}{233, 254, 187}
\definecolor{m_orange}{RGB}{255, 212, 121}
\definecolor{m_red}{RGB}{255, 190, 188}
\definecolor{m_violet}{RGB}{215, 131, 255}
\definecolor{m_blue}{RGB}{186, 234, 255}
\definecolor{m_brown}{RGB}{255,212,120}
\definecolor{m_lightgreen}{RGB}{212,251,122}
\definecolor{notetext}{rgb}{0.7,0,0}

\definecolor{model_pink}{RGB}{235, 106, 164}
\definecolor{model_orange}{RGB}{250, 194, 122}
\definecolor{model_green}{RGB}{164, 210, 162}
\definecolor{model_gray}{RGB}{120, 120, 120}
\definecolor{model_yellow}{RGB}{251, 231, 171}
\definecolor{model_purple}{RGB}{190, 146, 211}

\usepackage{amssymb}%
\usepackage{pifont}%

\def\eg{\emph{e.g.}\@\xspace} 
\def\ie{\emph{i.e.}\@\xspace}

\newcommand{\circlenum}[1]{{\textcircled{\scriptsize{#1}}}}

\newcommand{\name}{Search3D}

\newcolumntype{Y}{>{\centering\arraybackslash}X}
\newcolumntype{Z}{>{\raggedleft\arraybackslash}X}

\usepackage{array}
\newcolumntype{P}[1]{>{\centering\arraybackslash}p{#1}}
\newcolumntype{M}[1]{>{\centering\arraybackslash}m{#1}}

\definecolor{darkblue}{RGB}{60, 82, 145}
\definecolor{kingblue}{RGB}{65, 105, 225}

\definecolor{background}{RGB}{226, 226, 226}
\definecolor{head}{RGB}{210, 78, 142}
\definecolor{rightArm}{RGB}{255, 176, 0}
\definecolor{leftArm}{RGB}{228, 162, 227}
\definecolor{rightForeArm}{RGB}{90, 64, 210}
\definecolor{leftForeArm}{RGB}{243, 232, 88}
\definecolor{rightHand}{RGB}{158, 143, 20}
\definecolor{leftHand}{RGB}{192, 100, 119}
\definecolor{torso}{RGB}{100, 143, 255}
\definecolor{hips}{RGB}{129, 103, 106}
\definecolor{rightUpLeg}{RGB}{243, 115, 68}
\definecolor{leftUpLeg}{RGB}{152, 200, 156}
\definecolor{rightLeg}{RGB}{149, 192, 228}
\definecolor{leftLeg}{RGB}{152, 78, 163}
\definecolor{rightFoot}{RGB}{129, 0, 50}
\definecolor{leftFoot}{RGB}{76, 134, 26}

\newlength\myheight
\newlength\mydepth
\settototalheight\myheight{Xygp}
\usepackage{tikz}
\newcommand*\circled[1]{{\footnotesize \tikz[baseline=(char.base)]{
            \node[shape=circle,draw,inner sep=2pt] (char) {#1};}}}

\definecolor{revisioncolor}{RGB}{0, 0, 255}

\begin{document}

\title{Search3D: Hierarchical Open-Vocabulary 3D Segmentation}

\author{Ayca Takmaz$^{1, 2 \; ^\mathrm{\dagger}}$,
Alexandros Delitzas$^{1}$, Robert W. Sumner$^{1}$, Francis Engelmann$^{1, 2, 3 ^*}$ \\ Johanna Wald$^{2 ^*}$, Federico Tombari$^{2}$%
\thanks{Manuscript received: September 26, 2024; Revised January 8, 2025; Accepted January 13, 2025.}%
\thanks{This paper was recommended for publication by Editor M. Vincze upon evaluation of the Associate Editor and Reviewers' comments. This work was supported in part by an SNSF PostDoc.Mobility Fellowship during Francis Engelmann's stay at Stanford University, in part by an Innosuisse grant (48727.1 IP-ICT), and in part by the Swiss National Science Foundation Advanced Grant 216260: ``Beyond Frozen Worlds: Capturing Functional 3D Digital Twins from the Real World''. Alexandros Delitzas is supported by the Max Planck ETH Center for Learning Systems (CLS).
} %
\thanks{$^{1}$Ayca Takmaz, Alexandros Delitzas, Robert W. Sumner and Francis Engelmann are with ETH Zurich, Switzerland.}
\thanks{$^{2}$Ayca Takmaz, Francis Engelmann and Federico Tombari are with Google Zurich, Switzerland, and Johanna Wald is with Google Munich, Germany.}
\thanks{$^{3}$Francis Engelmann is with Stanford University, USA.}
\thanks{$^\mathrm{\dagger}$ Work done at Google Zurich as an intern. $^*$ equal supervision.}
\thanks{Correspondence to {\footnotesize	\{ayca.takmaz@inf.ethz.ch\}}}
\thanks{Digital Object Identifier (DOI): see top of this page.}
}

\markboth{IEEE Robotics and Automation Letters. Preprint Version. Accepted January, 2025}
{Takmaz \MakeLowercase{\textit{et al.}}: Search3D: Hierarchical Open-Vocabulary 3D Segmentation}

\maketitle

\begin{abstract}
Open-vocabulary 3D segmentation enables exploration of 3D spaces using free-form text descriptions.
Existing methods for open-vocabulary 3D instance segmentation primarily focus on identifying \textit{object}-level instances but struggle with finer-grained scene entities such as \textit{object parts}, or regions described by generic \textit{attributes}.
In this work, we introduce \name{}, an approach to construct hierarchical open-vocabulary 3D scene representations, enabling 3D search at multiple levels of granularity:
fine-grained object parts, entire objects, or regions described by attributes like materials.
Unlike prior methods, \name{} shifts towards a more flexible open-vocabulary 3D search paradigm, moving beyond explicit object-centric queries.
For systematic evaluation, we further contribute a scene-scale open-vocabulary 3D part segmentation benchmark based on MultiScan, along with a set of open-vocabulary fine-grained part annotations on ScanNet++.
\name{} outperforms baselines in scene-scale open-vocabulary 3D part segmentation, while maintaining strong performance in segmenting 3D objects and materials.
Our project page is 
\textit{\hypersetup{urlcolor=black}\href{https://search3d-segmentation.github.io}{search3d-segmentation.github.io}}.
\end{abstract}


\IEEEpeerreviewmaketitle

\section{Introduction}
\label{sec:intro}

\IEEEPARstart{E}{xtracting} semantic meaning from 3D scenes has traditionally relied on identifying a fixed set of pre-defined classes. For this purpose, most 3D segmentation methods~\cite{mask3d, 3d-mpa, minkowski} are trained on annotated datasets, limiting their capabilities to closed-set segmentation. While effective for these pre-defined classes, such approaches struggle to generalize to novel classes. However, personal and assistive robotics systems must operate in diverse, unknown environments and handle tasks of varying complexity, requiring the ability to handle unseen classes.
This calls for methods that can adapt to new tasks and environments, especially in human-centric spaces, which are inherently complex and composed of fine-grained elements critical to scene interaction. While identifying novel classes is already challenging, many interactive robotics applications \cite{lemke2024spotcompose} require identifying not only objects, but also their finer-grained \textit{components}~\cite{scenefun3d}, such as elevator buttons, light switches or chair seats. Further, attributes may vary across parts of an object, \eg{}, the seat of a chair may be leather while its legs are wooden. Such distinctions are crucial for tasks like cleaning, where robots must handle different materials with appropriate cleaning agents.
A purely object-centric understanding fails to provide this level of detail.
Ultimately, systems designed for such real-world interactions must be able to identify scene entities based on flexible and user-defined descriptions.

Open-vocabulary 3D segmentation methods~\cite{openscene, openmask3d, nguyen2023open3dis, conceptfusion, conceptgraphs, opennerf, opencity3d2025} have recently attracted growing interest \cite{opensun3d} and demonstrated promising results. These open-vocabulary methods can be grouped based on the underlying scene representation used to aggregate the features: a) instance-level \textit{object-centric} representations such as OpenMask3D~\cite{openmask3d} or Open3DIS\cite{nguyen2023open3dis}, or b) semantics-oriented point-level representations such as OpenScene~\cite{openscene}, OpenNeRF~\cite{opennerf} or ConceptFusion~\cite{conceptfusion}.

\begin{center}
\begin{figure}[t]
\centering
\includegraphics[width=0.48\textwidth]{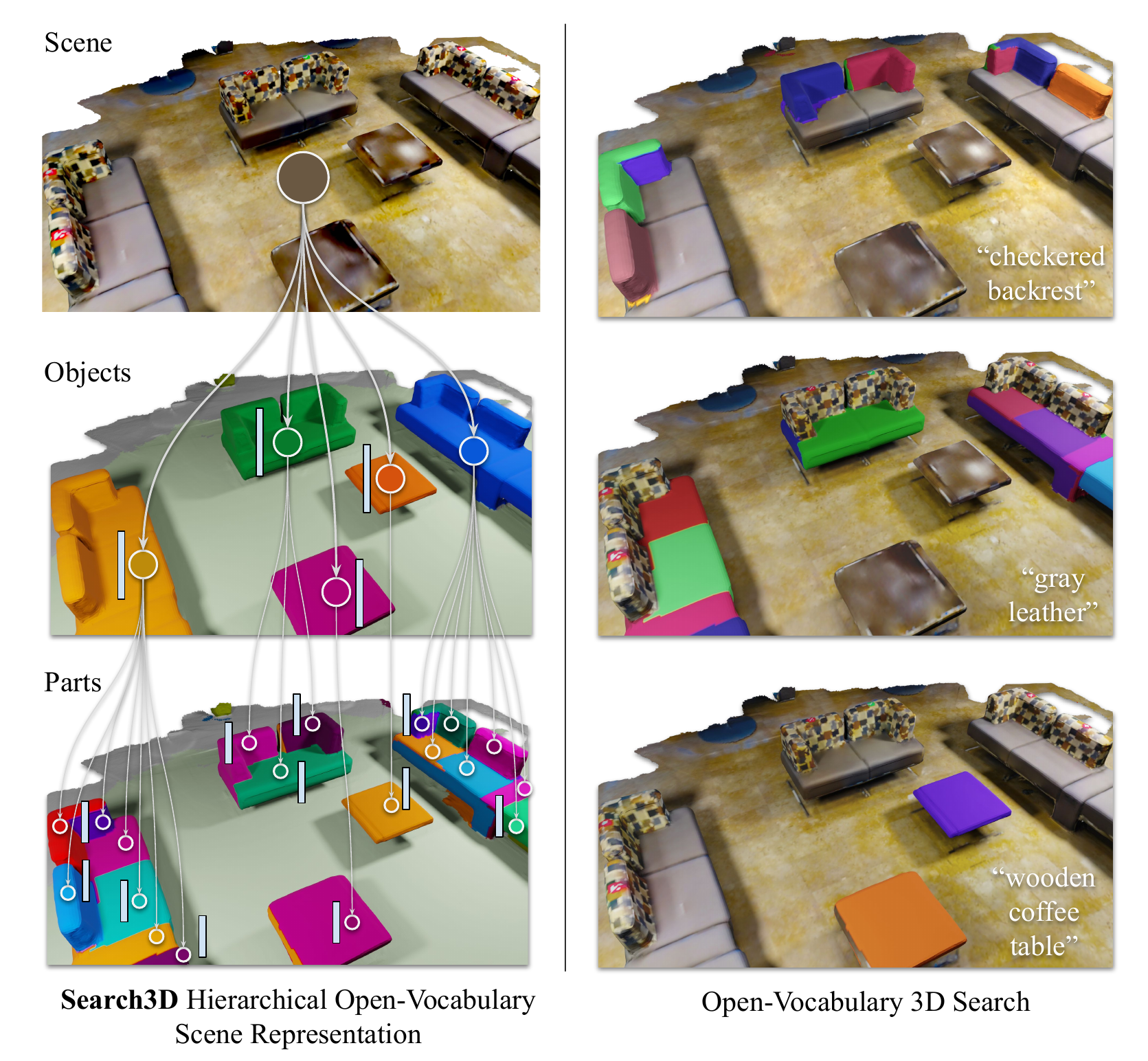}
\caption{
We propose Search3D, a method for open-vocabulary 3D search at multiple levels of granularity.
From posed RGB-D images and reconstructed geometry, we build a hierarchical scene representation with embedded features for objects, and finer-grained parts \emph{(left)}.
This enables searching across objects, parts, and attributes matching any given user query \emph{(right)}.}
\vspace{-15pt}
\label{fig:teaser}
\end{figure}
\end{center}

Object-centric open-vocabulary 3D segmentation methods typically first extract a set of class-agnostic 3D object instance masks and then compute a feature representation per object, represented in the joint vision-language embedding space of models such as CLIP~\cite{clip}. These methods are characterized by compact scene representations and are well-suited for directly segmenting object instances that match a given open-ended query. They are however not designed to identify scene entities of varying levels of granularity, \eg{}, ``seat of a chair”.

In contrast, other 3D open-vocabulary segmentation methods such as OpenScene~\cite{openscene}, OpenNeRF~\cite{opennerf} and ConceptFusion~\cite{conceptfusion} build per-point representations that aggregate features for each 3D point, resulting in a more fine-grained understanding of the scene.
However, storing these per-point features is memory-intensive, they are inherently noisy,
and they lack instance-level information -- a critical requirement for real-world applications in which a robot must identify the specific object to interact with among multiple instances \cite{lemke2024spotcompose, zurbrugg2024icgnet}.
Finally, the least obvious limitation is derived from the way these models compute the point-level features:
Although the projected open-vocabulary features are fine-grained at the level of the geometrical scene representation, the intermediate 2D feature backbones these methods use lack the detailed level of semantic meaning and are biased towards an object-level understanding.
Consequently, these methods often cannot robustly identify object parts and fine-grained elements, or address queries that describe areas spanning multiple regions of the scene, \eg, material segmentation.

In light of these limitations, we advocate for fine-grained open-vocabulary 3D segmentation to encompass a broader array of scene elements. An ideal open-vocabulary 3D segmentation method should robustly segment not only long-tail objects (``Nerf gun''), but also object parts (``chair backrest'') and queries that span multiple regions (``wooden''), while separating instances when necessary. This goes beyond the capabilities of existing methods. Our goal is to develop a method that moves beyond a strictly \textit{object-centric} query paradigm, and to move closer towards more flexible open-vocabulary 3D search capabilities.

Inspired by this vision, we propose \name{}, a hierarchical open-vocabulary 3D instance segmentation method based on a tree-structured hierarchical scene graph.
\name{} segments 3D scene entities from an arbitrary textual query, whether targeting object instances (level 1) or object parts (level 2), as shown in Figure~\ref{fig:teaser}.
To achieve this, we construct a tree representation where nodes represent scenes, objects and part-entities. For each object and part node, we compute open-vocabulary features enabling 3D segmentation across all levels.

To evaluate our method, we introduce a novel evaluation suite for open-vocabulary scene-scale 3D part segmentation based on MultiScan~\cite{multiscan}. Additionally, we perform experiments using hierarchical annotations for selected ScanNet++~\cite{scannetpp} scenes. Our method outperforms baselines for 3D open-vocabulary segmentation of object instances, as well as object parts (MultiScan, ScanNet++), and is able to segment the scene beyond instances, \eg{}, material segmentation (3RScan \cite{rio}). To summarize our key contributions:

\begin{itemize}
    \item We propose a hierarchical open-vocabulary 3D segmentation method capable of segmenting both entire objects and their parts given arbitrary textual queries, by aggregating features anchored to different granularity levels in a hierarchical tree structure.
    \item We introduce a benchmark for \textit{open-vocabulary scene-scale 3D part segmentation} by adapting MultiScan~\cite{multiscan} dataset for \textit{open-vocabulary} 3D part segmentation. This data will be released publicly through our project page.
    \item We contribute open-vocabulary hierarchical part annotations for a selection of ScanNet++~\cite{scannetpp} scenes.
    \item Our approach outperforms baselines on open-vocabulary 3D segmentation of object instances, part-level tasks, and scene-scale tasks such as material segmentation.
\end{itemize}

\section{Related work}
\label{sec:related-work}

\subsection{Open-Vocabulary \& Hierarchical 3D Scene Understanding}
Existing open-vocabulary 3D scene understanding methods typically focus on either object-level segmentation or point-level semantic segmentation.
Object-centric methods like OpenMask3D~\cite{openmask3d} and Open3DIS~\cite{nguyen2023open3dis} are limited to object-level segmentation and cannot handle varying levels of granularity.
In contrast, point-level methods such as OpenScene~\cite{openscene} and ConceptFusion~\cite{conceptfusion}, along with open-vocabulary implicit methods such as LeRF~\cite{lerf} and OpenNeRF~\cite{opennerf} provide sufficiently detailed features for fine-grained segmentation but lack structured, hierarchical representations.
Hierarchical querying is partially supported by N2F2~\cite{n2f2}, a 3D Gaussian splatting-based method embedding hierarchical features within a neural scene representation.
However, it does not enable \textit{explicit} part-\textit{instance} queries. 
Human3D~\cite{human3d} provides a hierarchical decomposition of humans into body parts but is limited to human-specific segmentation and lacks open-vocabulary capabilities.
GARField~\cite{garfield2024} represents scene elements with an affinity field for multi-granularity grouping but lacks language-guided querying capabilities.
Segment3D~\cite{Huang2023Segment3D} supports segmenting 3D scenes at varying granularity, akin to SAM for 2D image segmentation. SceneFun3D~\cite{scenefun3d} focuses on segmenting functional, interactable sub-parts in 3D environments given task-oriented language queries.
AGILE3D~\cite{yue2023agile3d} takes an entirely different approach, allowing users to segment arbitrary objects, parts, or elements through click-based interactions. Nevertheless, these works~\cite{Huang2023Segment3D, scenefun3d, yue2023agile3d} do not provide an explicit hierarchical structure.

\vspace{-20pt}
\begin{center}
\begin{figure*}[t]
\vspace{5px}
\centering
     \includegraphics[width=0.95\textwidth]{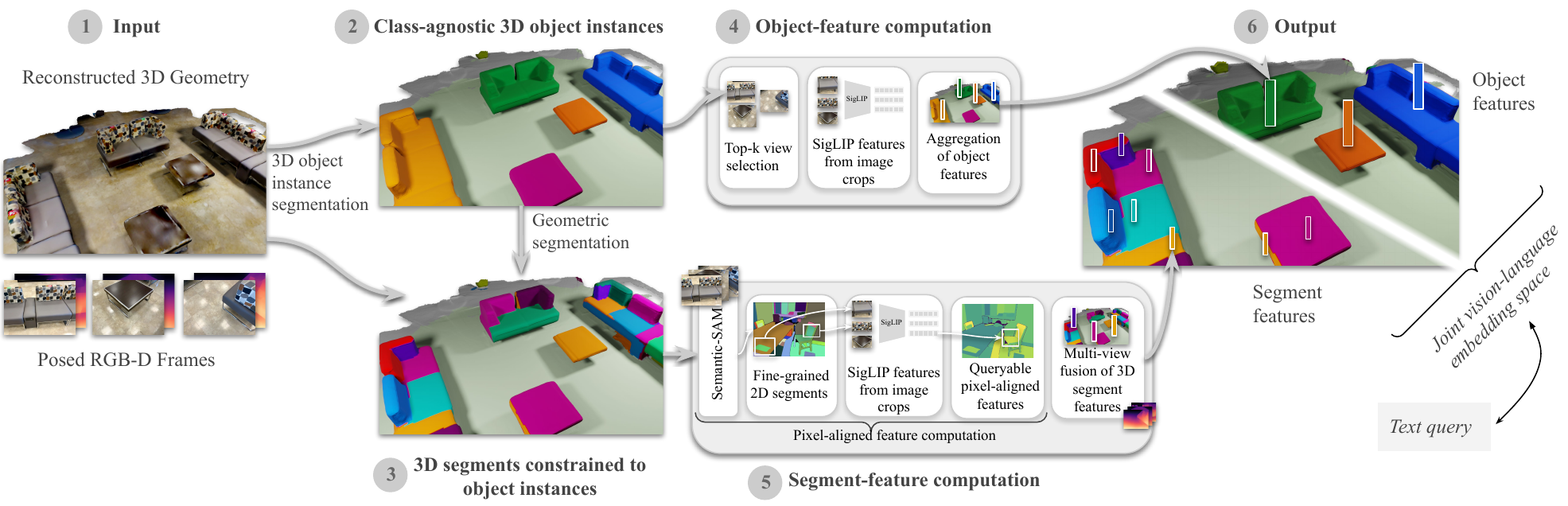}
\caption{\textbf{\name{} overview}: \circlenum{1} The inputs of our approach are posed RGB-D images of a 3D indoor scene along with its reconstructed 3D geometry. \circlenum{2} computes class-agnostic 3D instances which are passed to a geometric segmentation method \circlenum{3}, yielding a hierarchical 3D scene representation. In steps \circlenum{4} and \circlenum{5}, feature vectors are obtained for each object and segment. The hierarchical output representation \circlenum{6} is queryable with open-vocabulary features for objects and their corresponding parts enabling search in 3D via arbitrary text queries.}
\label{fig:overview_figure}
\vspace{-10pt}
\end{figure*}
\end{center}

Recent works~\cite{chen2024reason3D, partslip, satr, ma20243d} have explored 3D part segmentation in an open-vocabulary setting.
However, these methods are limited to single-object representations, focusing on part segmentation within individual object point clouds rather than handling scene-scale inputs.
Additionally, language-guided segmentation methods relying on GLIP~\cite{glip} require queries to be defined \textit{during} the construction of the representation, necessitating reprocessing with the original images for each new query. 
In contrast, our method builds an intermediate hierarchical feature representation, enabling open-vocabulary segmentation without prior query knowledge or storing the input images.
This allows for efficient querying during inference, making it well suited for real-world applications.

Other methods, such as HOV-SG~\cite{werby23hovsg} and CLIO~\cite{clio} use hierarchical open-vocabulary 3D scene graphs for robotic navigation.
However, these approaches operate at higher levels of abstraction—floor, room, region, or object—while our method extends the hierarchy to include finer-grained decomposition of objects into their smaller parts.

\subsection{VLMs and Open-Vocabulary Image Segmentation}
Large-scale vision-language models (VLMs) like CLIP~\cite{clip}, SigLIP~\cite{siglip}, and SILC~\cite{silc} provide a joint embedding space for image and text encoders.
While effective for tasks like image classification, their global per-image embeddings are not suited for pixel-aligned segmentation tasks.
To address this, methods OpenSeg~\cite{openseg}, LSeg~\cite{lseg} and others \cite{ovseg, maskclip, xdecoder, cat-seg, clip-seg} offer pixel-aligned representations, associating each pixel with an embedding vector. However, their training on full-object masks limits their ability to segment fine-grained entities like object parts.
Recent works~\cite{hipie, ov-parts, vlpart} tackle open-vocabulary part segmentation but rely on text queries as inputs to the segmentation network, lacking an explicit intermediate feature representation. This makes them unsuitable for building 3D open-vocabulary representations with the desired part-segmentation capabilities.
\section{Method}
\label{sec:method}

We introduce a novel hierarchical 3D scene representation enabling open-vocabulary segmentation for scene entities at multiple granularities, including objects and their parts. This representation is built upon 3D scenes reconstructed using posed RGB-D image sequences, as shown in Fig.~\ref{fig:overview_figure} \circled{1}, and addresses two key challenges:
\begin{enumerate}
    \item Representing scene entities at both object and part levels, Fig.~\ref{fig:overview_figure} \circled{2} and \circled{3}, discussed in Sec.~\ref{section:method_representation}.
    \item Computing open-vocabulary features for the scene representation, Fig.~\ref{fig:overview_figure}  \circled{4} and  \circled{5} described in Sec.~\ref{section:method_semantic_features}.
\end{enumerate}

\subsection{Hierarchical 3D Scene Representation}
\label{section:method_representation}

To capture both whole objects and their finer components, we construct a hierarchical scene representation as a tree structure (Fig.~\ref{fig:teaser}). The root node represents the \textit{scene}, comprising class-agnostic \textit{object} instances, which are further subdivided into smaller object components, \eg{}, object \textit{parts}. 

Our approach starts with an object-level mask proposal module, $\mathcal{F}_{obj}$,
built on a transformer-based Mask3D backbone~\cite{mask3d} pretrained on ScanNet200~\cite{scannet200}.
This module extracts class-agnostic object-level instances from the reconstructed 3D scene geometry.
Given the 3D scene $P_{scene} \in \mathbb{R}^{N\times3}$ where $N$ is the number of points, it outputs $M$ binary instance masks $\mathbf{M} = \mathcal{F}_{obj}(P_{scene}) = \{\mathbf{m}_1^{3D}, \mathbf{m}_2^{3D} ..., \mathbf{m}_M^{3D}\}$.
These masks represent the object nodes at the first level of our hierarchical scene representation.

The second stage of our method is the part-level segmentation module, $\mathcal{F}_{seg}$, which refines object instances into more granular segments $\mathbf{S}$. For each object $m$, we apply an instance-aware geometric over-segmentation technique which computes a set of segments $\mathbf{S}_m$ such that $\mathbf{S}_m = \mathcal{F}_{part}(P_{obj,m}) = \{\mathbf{s}_1^{3D}, \mathbf{s}_2^{3D} ..., \mathbf{s}_S^{3D}\}$, where $P_{obj,m} \in \mathbb{R}^{N_m \times 3}$ represents the 3D points that correspond to the predicted object mask $\mathbf{m}_m^{3D}$.
The segmentation module is a 3D adaptation of the graph-cut segmentation algorithm used in \cite{scannet}, originally proposed by \cite{felzenszwalb}.
Instead of segmenting the entire scene using this geometric segmentation approach,
we use the previously computed object masks $\mathbf{M}$ to segment each instance individually.
This ensures that the resulting segments remain within the boundaries of a single object, preserving the hierarchical tree structure.
Further, it guarantees that each segment contains points from only one object preventing overlap across masks.

For geometric over-segmentation, we use a segmentation clustering threshold of 0.05, which adaptively controls the merging of regions based on edge weights, and we require at least 100 vertices per segment to prevent over-fragmentation.

So far, we have computed scene entities hierarchically using a geometric representation of 3D object instances and their segments. While these masks capture spatial structures effectively,
they lack the semantic information needed for open-vocabulary 3D search.
Next, we detail how we enrich these scene entities with open-vocabulary features, enabling flexible 3D segmentation from free-form text queries.
\begin{figure}[t]
\centering
\vspace{5px}
\begin{overpic}[width=0.44\textwidth]{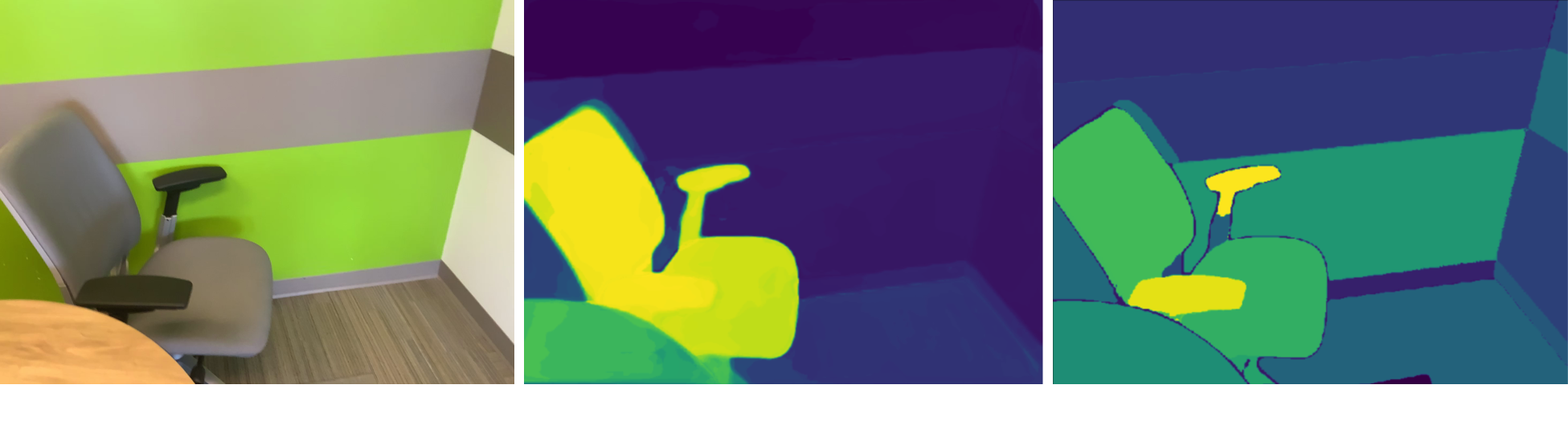}
\put(3.0,0){{\small \colorbox{white}{\textit{``black armrests}}}}
\put(6.8,-3.5){{\small \colorbox{white}{\textit{of the chair''}}}}
\put(39,0){{\small \colorbox{white}{OpenSeg \cite{openseg}}}}
\put(68,0){{ \small  \colorbox{white}{\name{} (Ours)}}}
\end{overpic}
\vspace{10px}
\caption{\textbf{Pixel-level features.} OpenSeg~\cite{openseg}, used in OpenScene, has a limited understanding of finer-grained object parts in the scene. We propose to obtain pixel-aligned features by combining Semantic-SAM segments~\cite{semanticsam} and SigLIP~\cite{siglip}, enabling fine-grained localization of concepts such as object parts and materials. Bright yellow means higher similarity to the text query.}
\vspace{-12px}
\label{fig:openseg_vs_ours}
\end{figure}

\subsection{Bringing Semantic Meaning to 3D Scenes}
\label{section:method_semantic_features}
To enable querying of scene entities across hierarchical levels, both object and part-level features are co-embedded in a shared embedding space using the SigLIP~\cite{siglip} VLM.
Building on the hierarchical 3D scene representation from Sec.~\ref{section:method_representation}, semantic features are  explicitly computed at two levels: \textit{objects} and \textit{part segments} as illustrated in Fig.~\ref{fig:overview_figure} \circled{4} and \circled{5}.

\noindent\textbf{Object-features} \circled{4} are extracted using a method inspired by \cite{openmask3d} and \cite{nguyen2023open3dis}, leveraging class-agnostic object masks to identify optimal views for semantic feature extraction.
These views are selected based on the projection characteristics of the object masks $\mathbf{M}$ initially generated by the object predictor \circled{4}.
For each 3D object proposal and camera pose, the visibility ratio is computed by projecting the object's points onto the camera image.
These visibility scores are then ranked in descending order, and the top-$K$ views ($K=5$) with the highest visibility ratios and therefore with minimal occlusion are selected.
To streamline processing, we subsample the RGB sequence by selecting every 5th frame, following \cite{openmask3d}.

For each selected view of an object, we first crop the image around a tight 2D bounding box that encapsulates the projected object points.
To gradually incorporate more scene context, we perform multi-scale cropping by extending the bounding box by a ratio of $k_{exp} = 0.2$ for $L$ steps, producing $L$ crops per view ($L=3$). This results in $K \cdot L$ image crops per object.
These crops are encoded into image embedding vectors of dimension $D=1152$ using the SigLIP~\cite{siglip} image encoder (So-400m).
The final feature vector \circled{6} is obtained by average pooling embeddings across all multi-view crops.

\noindent\textbf{Segment-features} \circled{5}, particularly for smaller entities such as object parts, are more challenging to extract.
While technically feasible to adapt the object-feature computation -- selecting optimal views for image crops -- our experiments reveal that this approach yields less informative features for segments, which are typically much smaller in scale.
Pixel-aligned VLMs like OpenSeg~\cite{openseg} offer potential for capturing pixel-level details but, as illustrated in Fig.~\ref{fig:openseg_vs_ours}, they are biased towards object-level understanding and often fail to represent the fine granularity required for smaller object parts.

To address this challenge, we propose a method to extract pixel-aligned features capable of representing finer-grained scene entities.
Using pixel-to-3D point mapping, features are directly aggregated for each predicted 3D part-segment (from \circled{3}) individually, as illustrated in \circled{5} of Fig.~\ref{fig:overview_figure}.
First, we apply the automatic mask generator from Semantic-SAM~\cite{semanticsam} to \textit{all} images in our RGB sequence, specifying the three highest granularity levels to consistently generate 2D segments representing smaller object parts.
Following a cropping strategy similar to the one for object-feature computation, we expand the tightest fitting 2D bounding box by a factor of $k_{exp}=0.1$ to obtain image crops of the fine-grained segments. These 2D segment crops are then passed through the SigLIP~\cite{siglip} image encoder, producing feature vectors of dimension $D$ for each segment.
Since our 2D segments are non-overlapping, the computed segment feature vectors are assigned to all pixels within each segment, producing a queryable pixel-aligned feature representation with shape $H \times W \times D$, where $H$ and $W$ represent the height and width of the image and $D$ is the feature dimensionality.
Finally, these multi-view features are fused at the 3D segment-level through average pooling. This step leverages the camera poses and scene geometry associated with each RGB frame to align the pixel-level features with the corresponding 3D segments (as extracted in \circled{3}).

Since the initial 3D segmentation is a geometric over-segmentation of each object \circled{3}, some parts may be split into multiple segments, even if they belong to the same component (\eg{}, the front and back parts of a chair's backrest).
To address this, we perform a semantically-informed merging of part-segments at the final stage.
For each 3D segment, neighboring segments within the same object that exhibit similar features are identified and merged based on two constraints:\\
\textbf{1) Proximity:} The closest distance between points in the segments is below a threshold $thr_{dist}$.\\
\textbf{2) Feature similarity:} The cosine similarity of their feature vectors exceeds a threshold $thr_{feat}$.\\
Pairs of segments satisfying both conditions are iteratively merged until no more candidates meet the criteria.
We set $thr_{dist}=0.07$ and $thr_{feat}=0.13$.
Once all candidate segment pairs are merged, the new 3D segment becomes the union of the merged components, and its feature vector is the average of all contributing features. This refinement updates the 3D segments in the hierarchical scene representation, reflecting the semantic merging.

\noindent\textbf{Hierarchical open-vocabulary 3D search.}
Our hierarchical feature representation enables 3D search across multiple levels of granularity, enhancing open-vocabulary 3D segmentation. When querying for a specific part of an object, we leverage both part- and object-level semantic information encoded in the hierarchical representation. 
Given an input query such as ``seat of a chair'', we first encode the query using the SigLIP text encoder to obtain an embedding vector $e_{txt} \in \mathcal{R}^{D}$. 
In our hierarchical representation, each object and segment is associated with feature vectors: $e_{obj}$ for the object node and $e_{seg}$ for the part segment node.
For any pair of features $(e_{obj}, e_{seg})$ -- where $e_{obj}$ represents the parent object feature and $e_{seg}$ represents the child part feature -- we compute the overall similarity with the query text embedding $e_{txt}$ as follows: $sim_{query} = avg(cos\_sim(e_{txt}, e_{obj}), cos\_sim(e_{txt}, e_{seg}))$. Here, $cos\_sim$ denotes the cosine similarity between L2-normalized embedding vectors. This average similarity score captures the combined relevance of object-level and segment-level features to the query.

By leveraging this hierarchical approach, we can effectively capture the desired object parts, even when the query targets a specific segment of a larger object.
This method enables accurate identification and retrieval of both individual parts and their broader contextual components within 3D scenes. Such flexibility ensures more precise and contextually relevant results for fine-grained 3D segmentation tasks.
\section{Data}
\label{sec:data}

In this work, we aim to extend open-vocabulary 3D segmentation capabilities across different granularity levels. To this end, we evaluate our method's ability to perform scene-scale 3D part-level instance segmentation guided by open-vocabulary descriptions. Comprehensive evaluation requires a dataset with scene-scale annotations that ideally captures the object-part hierarchy.

\vspace{-10pt}
\subsection{Open-Vocabulary 3D Part Segmentation on MultiScan} \label{subsec:data-multiscan}
As mentioned earlier, the MultiScan~\cite{multiscan} dataset is the only available resource that includes both scene-scale \textit{object} and \textit{part} instance annotations.
It provides essential assets such as RGB-D sequences, calibrated camera data, and 3D surface meshes.
Crucially, it includes part- and object-level semantic labels that preserve the scene-object-part hierarchy, making it highly suitable for our work.

The MultiScan dataset was originally annotated with 419 fine-grained categories, later grouped into coarser category sets.
For 3D object-level instance segmentation, the original benchmark focuses on 17 common object categories.
However, for 3D part-instance segmentation, it only includes 5 part-semantic categories: \textit{static, door, drawer, window, lid}. 
While these 5 categories is meaningful for MultiScan's original focus on articulated part segmentation, they are insufficient for the \textit{open-vocabulary} scenarios we aim to address.
To enable a broader evaluation, we analyzed the existing MultiScan annotations and identified a larger set of object and part categories suitable for open-vocabulary evaluation.
The adapted dataset we release, based on existing fine-grained annotations from MultiScan, includes 155 object and 15 part categories.

Another key consideration for open-vocabulary part segmentation is the meaningfulness of part names at the scene scale.
Existing part annotations in MultiScan specify only the semantic category of the part, such as ``door''.
However, this can be problematic when performing and evaluating open-vocabulary part segmentation based solely on the part category name.
For example, a "desk" object may have a part labeled as "door," which, without additional context, could cause confusion about the intended scene entity. Even humans might incorrectly associate it with the typical meaning of a "door."
To mitigate this, we recognize that \textit{(object, part)} pairs are generally more informative for identifying part-level entities in the scene, such as the ``seat'' of a ``chair'', or the ``door'' of the ``cabinet''. Based on this insight, we extracted a set of 47 joint object-part labels from the MultiScan dataset, consisting of these more informative (object, part) pairs.

\subsection{Fine-grained Part Annotations on ScanNet++} \label{subsec:data-scannetpp}

\begin{center}
\begin{figure}[!t]
\centering
\vspace{5px}
  \begin{overpic}[width=0.49\textwidth]{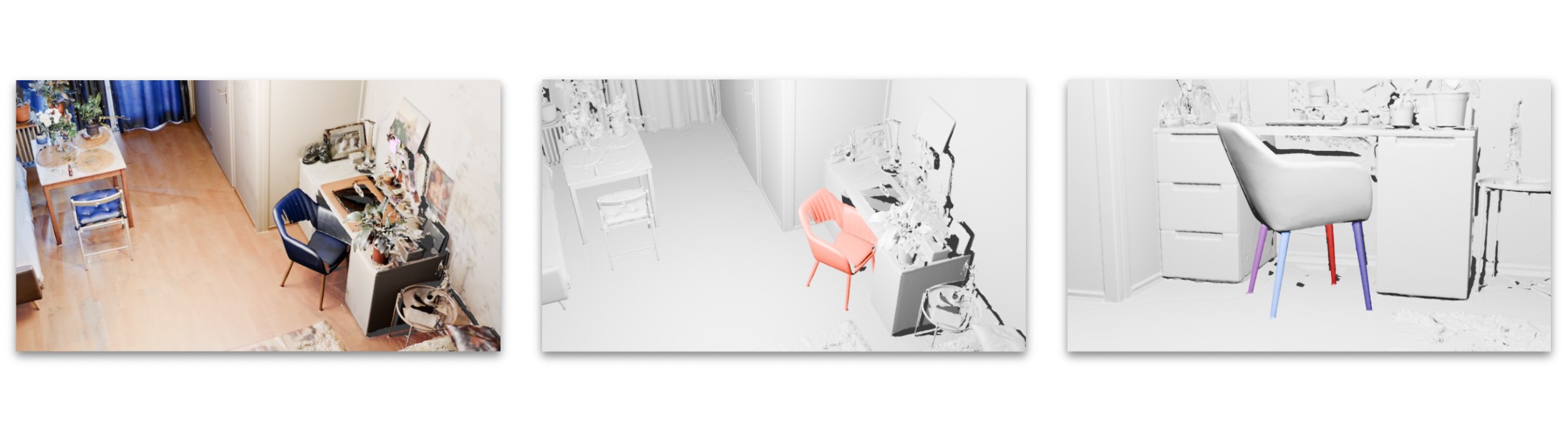}%
    \put(12,24){{\small \colorbox{white}{\textbf{Scene}}}}
    \put(44.5,24){{\small \colorbox{white}{\textbf{Object}}}}
    \put(70.5,24){{\small \colorbox{white}{\textbf{Annotated Parts}}}}
    
    \put(37.5,1.5){{\small   \colorbox{white}{\textit{``blue armchair''}}}}
    \put(77.5,1.5){{ \small  \colorbox{white}{\textit{``legs''}}}}
\end{overpic}

\vspace{-5px}     
\caption{\textbf{An example from our hierarchical object and part annotations on a selection of ScanNet++ \cite{scannetpp} scenes.} 
}

\label{fig:our_annotations}
\end{figure}
\end{center}

To evaluate our method on fine-grained object and part segmentation, we provide an additional evaluation dataset with annotations on laser scans (Fig.~\ref{fig:our_annotations}).
As highlighted in SceneFun3D~\cite{scenefun3d} and ScanNet++~\cite{scannetpp}, laser scans capture finer 3D geometry details of object parts in indoor environments -- details often missing in datasets captured with commodity devices (\eg{}, iPhone), such as MultiScan~\cite{multiscan}.
To address this gap, our dataset includes 14 object and 20 part annotations across 8 ScanNet++~\cite{scannetpp} scenes, along with open-vocabulary text descriptions.
Using the SceneFun3D annotation tool~\cite{scenefun3d}, we performed fine-grained semantic annotation on high-resolution point clouds, and extended it to incorporate object-part hierarchy information.

\vspace{-30pt}
\section{Experiments}
\label{sec:experiments}
To assess our method's ability to search and segment in 3D via arbitrary open-vocabulary queries, we evaluate it on three diverse tasks:
1) 3D part segmentation (Sec.~\ref{eval:parts}),
2) 3D object instance segmentation (Sec.~\ref{eval:instances}), and
3) 3D material segmentation (Sec.~\ref{eval:material}).
Additionally, we validate our design choices through corresponding ablation studies.

\subsection{3D Part Segmentation}
\label{eval:parts}
To evaluate our method's ability to handle queries beyond object-level descriptions, we introduce the task of scene-level 3D open-vocabulary part segmentation.
For this analysis, we use our adapted MultiScan~\cite{multiscan} scene-level part segmentation data (see Sec.~\ref{subsec:data-multiscan}), and the annotations we provide on the ScanNet++~\cite{scannetpp} dataset (see Sec.~\ref{subsec:data-scannetpp}).
In the part-level instance segmentation experiments, we report the Average Precision metric evaluated at 50\% (AP$_{50}$), 25\% (AP$_{25}$) overlap thresholds, and the average over the range of [0.5 : 0.95 : 0.05] (AP) following established benchmarks\cite{scannet}.

\begin{table}[!t]
\setlength{\tabcolsep}{17pt}
\centering
\begin{tabular}{l|c|l}
\toprule
     \textbf{Methods} & \textbf{Segments} & $\mathbf{AP}$\\
     \midrule
     OpenScene \cite{openscene} & Oracle  & $31.4$ \\
     OpenMask3D \cite{openmask3d} & Oracle & $35.7$\\
     \name{} (Ours)& Oracle  & $\mathbf{49.5}$ \textcolor{darkgreen}{\scriptsize(+\,$13.8$)} \\
     \bottomrule

\end{tabular}
\caption{\textbf{3D part feature quality evaluation on the MultiScan \cite{multiscan} dataset using GT part segments.} We conduct an oracle experiment using annotated GT part segments to aggregate features for OpenScene \cite{openscene}, OpenMask3D \cite{openmask3d} and \name{} (Ours) to measure the quality of the features computed from each method, when isolated from geometric segmentation performance.}
\label{tab:multiscan-oracle-joint}
\vspace{-10pt}
\end{table}

First, we evaluate the quality of our segment features for identifying object parts using an oracle mask experiment,
isolating feature quality from the effect of 3D geometric part segmentation quality.
In this analysis, all methods use ground-truth (GT) part segments:
For OpenScene, we aggregate per-point features for each GT 3D part segment, and for OpenMask3D we aggregate per-mask features for each GT part segment.
Tab.~\ref{tab:multiscan-oracle-joint} shows the results from this oracle experiment. It demonstrates the strong open-vocabulary part-segmentation performance of our segment-level features, with at least {+\,$13.8$} AP improvement over baseline methods.

\begin{figure}[t]
\centering
  \begin{overpic}[width=0.48\textwidth]{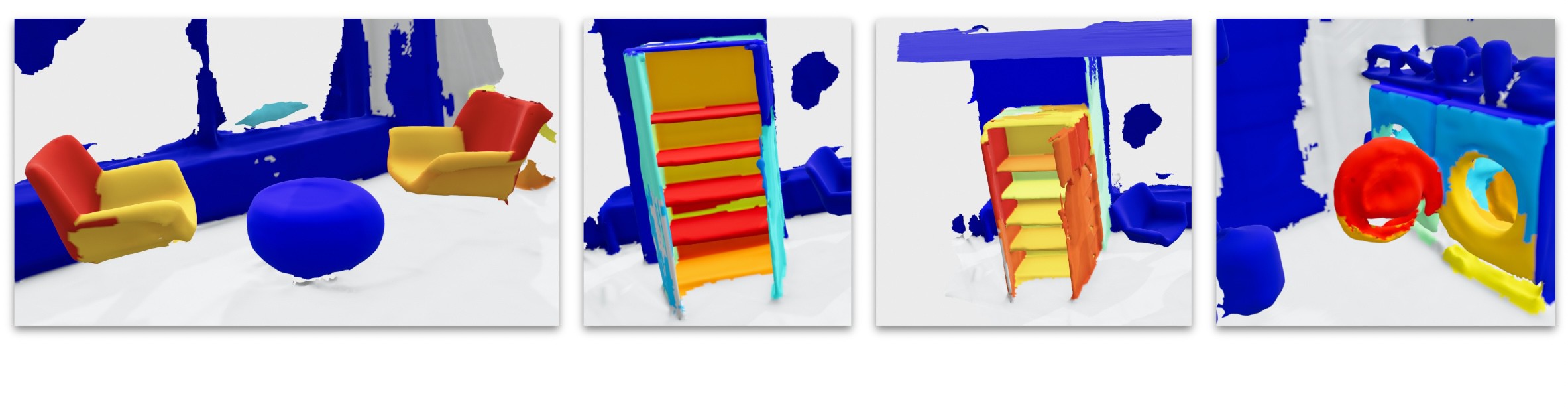}%
    \put(10,1){{\footnotesize \colorbox{white}{\textit{``back of the}}}}
    \put(35,1){{ \footnotesize \colorbox{white}{\textit{``shelf of the}}}}
    \put(55.5,1){{ \footnotesize \colorbox{white}{\textit{``sides of the}}}}
    \put(77.5,1){{ \footnotesize \colorbox{white}{\textit{``door of the}}}}
    \put(10,-3){{\footnotesize \colorbox{white}{\textit{   armchair''}}}}
    \put(35,-3){{ \footnotesize \colorbox{white}{\textit{   cupboard''}}}}
    \put(55.5,-3){{ \footnotesize \colorbox{white}{\textit{     cupboard''}}}}
    \put(75.5,-3){{ \footnotesize \colorbox{white}{\textit{washing machine''}}}}
\end{overpic}
\vspace{8pt}
\caption{{\textbf{Heatmaps showing response to text queries of \name{}}}.
Dark red means high similarity and dark blue means low similarity. 
}
\vspace{-5pt}
\label{fig:example_heatmaps}
\end{figure}

\begin{figure}[!t]
\centering
  \begin{overpic}[width=0.48\textwidth]{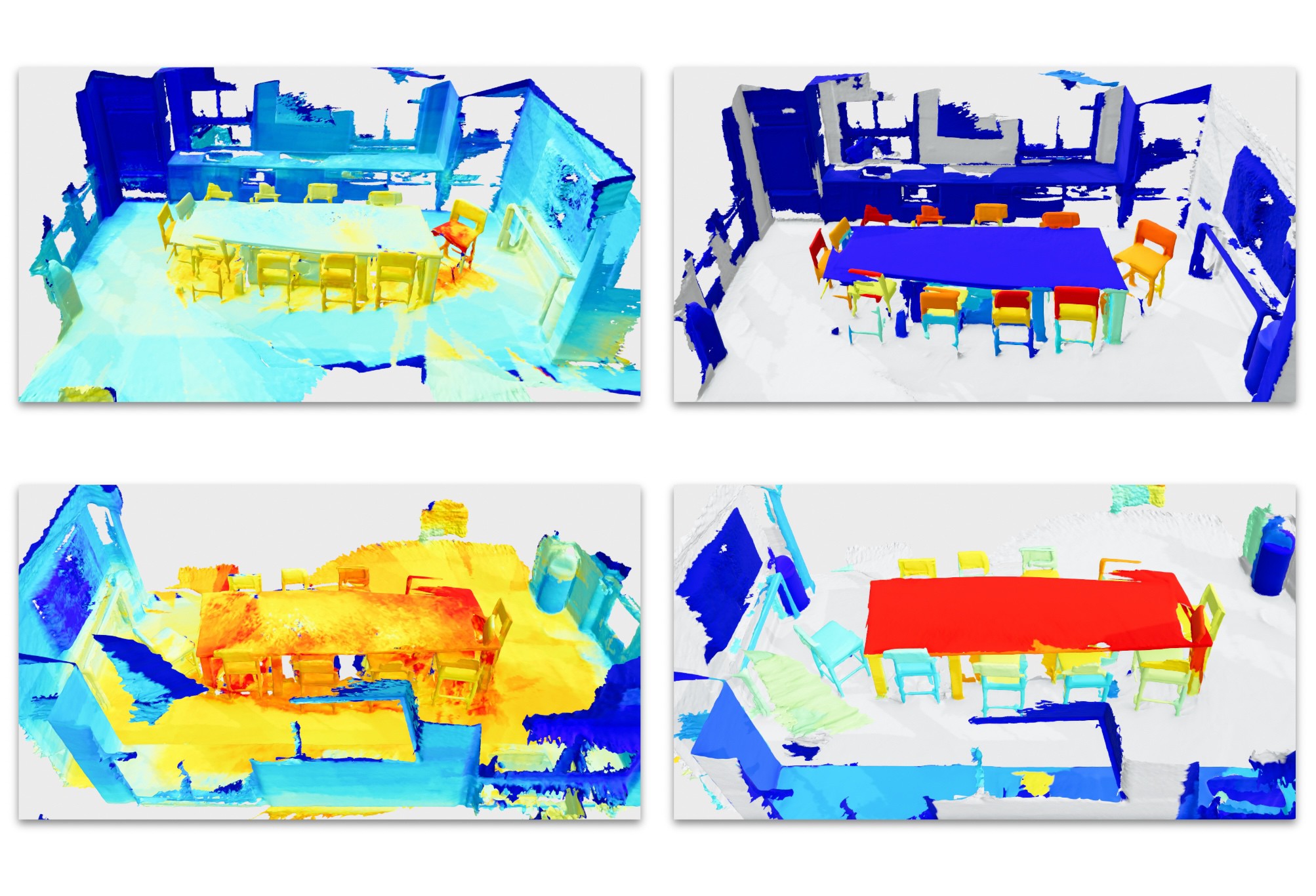}%
     \put(15,64.5){{\footnotesize \colorbox{white}{OpenScene \cite{openscene}}}}
      \put(63,64.5){{\footnotesize \colorbox{white}{ \name{} (Ours)}}}
    \put(30,34){{\footnotesize \colorbox{white}{\textbf{Query:} \textit{``backrest of a chair''}}}}
    \put(26.5,2){{\footnotesize \colorbox{white}{\textbf{Query:} \textit{``surface of the dinner table''}}}}
\end{overpic}
\vspace{-4pt}
\caption{\textbf{Similarity heatmaps between text queries and scene features.} We compare OpenScene \cite{openscene} per-point features with the segment features from our method. Dark red means high similarity and dark blue means low similarity. Our method shows highly localized understanding of object parts.
}
\vspace{-13pt}
\label{fig:comparison_heatmaps}
\end{figure}

After analyzing feature informativeness using \textit{ground-truth part} masks, we evaluate part segmentation performance using \textit{predicted} part masks on the adapted MultiScan dataset.
The results, presented in Tab.~\ref{tab:multiscan-joint-part-seg}, validate \name{}'s strong 3D part segmentation ability.
Fig.~\ref{fig:example_heatmaps} and \ref{fig:comparison_heatmaps} further show the improved part localization compared to methods like OpenScene, which relies on per-point feature representations.
Additionally, in Tab.~\ref{tab:scannetpp-ours-part}, we provide part-segmentation results on our ScanNet++ annotations (Sec.~\ref{subsec:data-scannetpp}). 

\begin{table}[!t]
\centering
\setlength{\tabcolsep}{5pt}
\resizebox{0.48\textwidth}{!}{
\begin{tabular}{l| c |c| c| c}
\toprule
     \textbf{Methods} & \textbf{Aggregation} & $\mathbf{AP}$  & $\mathbf{AP_{50}}$ & $\mathbf{AP_{25}}$\\
     \midrule
     (1) OpenScene \cite{openscene} & segments  & $3.2$  & $5.5$ & $13.7$ \\ %
     (2) OpenMask3D \cite{openmask3d} & objects  & $3.3$  & $6.1$ & $11.3$ \\
     (3) OpenMask3D \cite{openmask3d} & segments & $3.1$  & $6.2$ & $18.2$\\
    (4) GARField \cite{garfield2024} + Search3D &   segments  &   $3.5$  &   $8.9$ &  $20.5$ \\ %
      (5) GARField \cite{garfield2024} + Search3D &   seg. + hierarchy  &   $3.2$  &   $8.4$ &  $15.3$ \\ %
  
     (6) \name{} (Ours)& seg. + hierarchy & $\mathbf{7.9}$ & $\mathbf{14.5}$ & $\mathbf{31.5}$ \\
     
      & & \textcolor{darkgreen}{\scriptsize(+\,$4.6$)} & \textcolor{darkgreen}{\scriptsize(+\,$8.3$)} & \textcolor{darkgreen}{\scriptsize(+\,$13.3$)} \\
     \bottomrule

\end{tabular}
}
\caption{\textbf{3D Part Segmentation on MultiScan \cite{multiscan}.}
The queries combine object and part descriptions to perform open-vocabulary part retrieval. (1) uses 2D fused OpenSeg~\cite{openseg} feats., and per-point feats. are aggregated over part segments.
(2) uses the orig. object-level masks from OpenMask3D.
(3) is a stronger baseline adapted from (2) using segment-level aggregation.
(4) and (5) use object and part masks from GARField \cite{garfield2024} at scales $0.35$ and $0.1$ respectively and employ Search3D for feature computation using these masks. \name{} (6) uses all components, incl. hierarchical search.}
\label{tab:multiscan-joint-part-seg}
\end{table}

\begin{table}[!t]
\centering
\setlength{\tabcolsep}{15pt}
\begin{tabular}{l|c|c|c}
\toprule
     \textbf{Methods} & $\mathbf{AP}$ & $\mathbf{AP_{50}}$ & $\mathbf{AP_{25}}$\\
     \midrule
     OpenMask3D \cite{openmask3d}   & $5.2$ & $15.0$  & $18.1$ \\
     \name{} (Ours) & $\mathbf{17.0}$  & $\mathbf{32.4}$  & $\mathbf{38.3}$ \\
     \bottomrule
\end{tabular}
\caption{{3D part instance segmentation results on the set of annotations we provide on a selection of ScanNet++ \cite{scannetpp} scenes.}}
\label{tab:scannetpp-ours-part}
\end{table}%

\textbf{Ablation study.}
We analyze the impact of semantically informed segment merging and hierarchical search.
The results in Tab.~\ref{tab:multiscan-joint-part-seg-ablation} emphasize the importance of those components.
Semantically informed segment merging contributes \textcolor{darkgreen}{+\,$3.2$} AP$_{50}$,
while leveraging the scene hierarchy for part-level entity search adds \textcolor{darkgreen}{+\,$3.1$} AP$_{50}$.
Additionally, averaging the object-level and part-level similarity scores yields slightly better results than using the maximum of these scores. Overall, these enhancements lead to a combined improvement of \textcolor{darkgreen}{+\,$6.3$} AP$_{50}$.


\begin{table}[!t]
\centering
\setlength{\tabcolsep}{5pt}
\resizebox{0.48\textwidth}{!}{
\begin{tabular}{l|c|c|c|c|c|c}
\toprule
     \textbf{Methods} & \textbf{Seg.} &\textbf{ Merging} &\textbf{Hier. }  & &  & \\
      & \textbf{Aggr.} & &\textbf{search} & $\mathbf{AP}$  & $\mathbf{AP_{50}}$ & $\mathbf{AP_{25}}$\\
     \midrule
     (1) Ours& \checkmark & &&$4.7$  & $8.2$ & $17.6$ \\
     (2) Ours & \checkmark{}&\checkmark{}&  & $6.6$ & $11.4$ & $23.7$\\
     (3) Ours & \checkmark{}&\checkmark{} & \checkmark{} (max.)  & $7.5$ & $13.5$ & $28.4$ \\
     (4) Ours & \checkmark{}&\checkmark{} & \checkmark{} (avg.)  & $\mathbf{7.9}$ & $\mathbf{14.5}$ & $\mathbf{31.5}$ \\
      &&& & \textcolor{darkgreen}{\scriptsize(+\,$3.2$)} & \textcolor{darkgreen}{\scriptsize(+\,$6.3$)} & \textcolor{darkgreen}{\scriptsize(+\,$13.9$)} \\
     \bottomrule
\end{tabular}
}
\vspace{-2pt}
\caption{\textbf{Ablation study on \name{} components for 3D part segmentation on MultiScan \cite{multiscan}.} \textit{Merging} refers to post-processing and merging 3D segments based on their feature similarities. \textit{Hier. search} refers to measuring the overall similarity between text query and each segment using both object and part features.}
\label{tab:multiscan-joint-part-seg-ablation}
\vspace{-14pt}
\end{table}

\begin{table}[!b]
\centering
\setlength{\tabcolsep}{2pt}
\resizebox{0.48\textwidth}{!}{
\begin{tabular}{lccccccc}
\toprule

&  &   & & Head & Common & Tail \\
Model  & Masks & Img.  Feat.  & AP  & (AP) & (AP) & (AP) \\

\midrule
\multicolumn{6}{l}{\textcolor{black!60}{\textit{Closed-vocabulary, full sup.}}} \\[-0.2ex] %

\textcolor{black!100}{Mask3D \cite{mask3d}} & \textcolor{black!100}{Mask3D} & $\textcolor{black!30}{-}$ &$\textcolor{black!100}{26.9}$ & $\textcolor{black!100}{39.8}$ &$\textcolor{black!100}{21.7}$& $\textcolor{black!100}{17.9}$  \\ %

\arrayrulecolor{black!10}\midrule\arrayrulecolor{black}
\multicolumn{6}{l}{\textit{\textcolor{black!60}{Open-vocabulary (using 2D and 3D object mask predictors)}}} 
    \\[-0.2ex] 
Open3DIS \cite{nguyen2023open3dis} (2D\&3D) & SAM+ISBNet & CLIP & $23.7$  &$27.8$ & $21.2$ & $21.8$    \\

Open3DIS \cite{nguyen2023open3dis} (2D\&3D) & SAM+Mask3D & CLIP & $23.7$  &$26.4$ & $22.5$ & $21.9$    \\

\arrayrulecolor{black!10}\midrule\arrayrulecolor{black}
\multicolumn{6}{l}{\textit{\textcolor{black!60}{Open-vocabulary (using only 3D object mask predictors)}}}  & \\[-0.2ex] 

OpenScene \cite{openscene} (2D F.)  & Mask3D & OpenSeg & $11.7$ & $13.4$ &$11.6$ & $9.9$ \\ 

OpenMask3D \cite{openmask3d} & Mask3D & CLIP
&$15.4$ &$17.1$  &$14.1$ &$14.9$  \\ %

Open3DIS \cite{nguyen2023open3dis} (3D M.) & ISBNet & CLIP & $18.6$  &$24.7$ & $16.9$ & $13.3$  \\

Open3DIS \cite{nguyen2023open3dis} (3D M.) & Mask3D & CLIP & $18.9$  &$23.9$ & $17.4$ & $15.3$  \\

Ours (\name{}) & Mask3D  & CLIP &  $14.3$ & $16.1$ &$13.6$& $12.9$  \\ %

Ours (\name{}) & Mask3D  & SigLIP &  $\mathbf{23.0}$ & $\mathbf{26.3}$ &$\mathbf{21.2}$& $\mathbf{21.4}$  \\
\bottomrule
\end{tabular}
}
\caption{\textbf{3D Instance Segmentation on ScanNet200 val.}
Head, Common, and Tail are subsets of ScanNet200 classes, ordered by descending frequency \cite{scannet200}.
Our method, while capable of segmenting fine-grained scene entities such as object parts, thanks to its hierarchical representation also preserves strong open-vocabulary 3D object segmentation performance.} %
\label{tab:3d_inst_seg}
\end{table}

\subsection{3D Instance Segmentation}
\label{eval:instances}
Another key question is whether our method maintains strong open-vocabulary 3D instance segmentation performance while also being capable of segmenting part-level scene entities. To evaluate this, we compare our method with existing open-vocabulary 3D instance segmentation methods using the standard benchmark on ScanNet200~\cite{scannet200} in Tab.~\ref{tab:3d_inst_seg}, and additionally on MultiScan \cite{multiscan} in Tab.~\ref{tab:multiscan-obj}, using the AP metrics as defined earlier in Sec.~\ref{eval:parts}. As shown in Tab.~\ref{tab:3d_inst_seg}, our method has very strong 3D instance segmentation performance, outperforming other counterparts that rely solely on 3D masks for identifying object-level instances. These results also validate the effectiveness of using SigLIP~\cite{siglip} as a VLM for our method, resulting in strong gains compared to using CLIP~\cite{clip} to compute open-vocabulary features. Additionally, while GARField~\cite{garfield2024} provides a flexible grouping mechanism for segmenting 3D scenes at arbitrary granularities, the lack of an explicit objectness prior leads to inconsistent object segmentation. In contrast, our mask module can identify objects, which enables strong performance for both instance segmentation and part segmentation through the hierarchical search mechanism.

\begin{table}[!t]
\centering
\setlength{\tabcolsep}{15pt}
\resizebox{0.48\textwidth}{!}{
\begin{tabular}{lccc}
\toprule
 \textbf{Methods} & $\mathbf{AP}$ & $\mathbf{AP_{50}}$ & $\mathbf{AP_{25}}$\\
 \midrule
    GARField \cite{garfield2024} &   $2.4$ &   $5.6$ &   $9.6$ \\
 \color{black} OpenScene \cite{openscene}  &$9.0$ & $12.6$ & $16.7$ \\
 OpenMask3D \cite{openmask3d}   & $10.7$ & $15.7$  & $20.8$ \\
 \name{} (Ours) & $\mathbf{18.1}$  & $\mathbf{26.3}$  & $\mathbf{33.5}$ \\ %
 \bottomrule
\end{tabular}
}
\caption{
\textbf{3D object instance segmentation scores on MultiScan \cite{multiscan}.}
We compare with object-level features from our method.}
\label{tab:multiscan-obj}
\end{table}

\subsection{3D Material Segmentation}
\label{eval:material}
Next, we perform an analysis on 3D material segmentation task using the object-level material annotations from the 3RScan dataset~\cite{rio}. We use Intersection-over-Union (mIoU) and mean accuracy (Acc) to evaluate material class predictions obtained using query-similarity based assignments similar to the instance segmentation task. The results in Tab.~\ref{tab:3rscan-materials} highlight our method's ability to go beyond object semantics.

\begin{table}[!t]
\centering
\resizebox{0.49\textwidth}{!}{
\begin{tabular}{lcc}
\toprule
     \textbf{Methods} & \textbf{mIoU} & $\mathbf{Acc}$\\
     \midrule
     (1) MinkowskiNet \cite{minkowski} \textit{(fully supervised)} & $23.5$ & $30.6$ \\ 
\arrayrulecolor{black!10}\midrule\arrayrulecolor{black}
     \textit{Open-vocabulary, 3D distillation of features}  &  &   \\ 
     (2) OpenScene (3D distill) \cite{openscene} & $15.3$ & $26.4$ \\
     (3) OpenScene (2D/3D ensemble) \cite{openscene} & $20.1$ & $35.6$ \\
     \arrayrulecolor{black!10}\midrule\arrayrulecolor{black}
     \textit{Open-vocabulary, multi-view fusion} && \\ 
     (4) OpenScene (2D fusion) \cite{openscene} & $18.6$ & $31.9$ \\
     (5) \name{} (Ours) & $\mathbf{20.2}$   & $\mathbf{38.4}$  \\
     \bottomrule

\end{tabular}
}
\caption{
\textbf{3D material segmentation scores on 3RScan \cite{rio} using object-level material annotations from 3DSSG \cite{wald2020learning}.}
To assess the capabilities of our method on open-vocabulary segmentation with a focus on concepts other than object or part semantic categories, we present material segmentation results.}
\label{tab:3rscan-materials}
\vspace{-10px}
\end{table}

\subsection{Runtime Analysis}
In Tab.~\ref{tab:runtime_complexity}, we present a runtime analysis of our method. The construction of the open-vocabulary hierarchical representation \circlenum{1}-\circlenum{5} is performed offline. Once this representation is built, inference \circlenum{6}, \ie{}, 3D search based on user input queries can be performed at around 1-2 FPS.
\begin{table}[!t]
\centering
\setlength{\tabcolsep}{0.5pt}
\resizebox{0.49\textwidth}{!}{
\begin{tabular}{lcc}
\toprule

\textbf{Method component} & \textbf{Runtime} & \textbf{Proportional} \\
\midrule
\multicolumn{1}{l}{\textbf{\circlenum{2} \textit{3D Object Instance Segmentation}}} & (per-scene avg.)\\[-0.2ex]
Forward-pass of 3D instance seg. model & 0.55 s & - \\
Post-processing \& I/O & 18.43 s & $ T \propto M$\\
\arrayrulecolor{black!10}\midrule\arrayrulecolor{black}
Total (per-scene) & 18.98 s & - \\
\midrule
\multicolumn{3}{l}{\textbf{\circlenum{3} \textit{Geometric Segmentation for Part Segmentation}}} \\[-0.2ex]
Normal-based geometric segmentation & 4.33 s & - \\
Hierarchical tree formation \& I/O cost & 17.52 s & $ T \propto (M\cdot S)$\\
\arrayrulecolor{black!10}\midrule\arrayrulecolor{black}
Total (per-scene) & 21.85 s & - \\
\midrule
\multicolumn{1}{l}{\textbf{\circlenum{4} \textit{Object-Feature Computation}}} & (per-scene avg.)\\[-0.2ex]

Top-k view selection & 1.51 s & $ T \propto (  n_{\mathrm{frames}} \cdot M)$  \\
Pre-computation of point projections & 32.00 s & $ T \propto (  n_{\mathrm{frames}} \cdot M)$ \\
Multi-level image crops & 2.46 s & $ T \propto M$ \\
SigLIP features from image crops & 215.62 s & $ T \propto M$ \\
Aggregation of object-features & 3 ms & - \\
I/O overhead & 15.76 s & - \\
\arrayrulecolor{black!10}\midrule\arrayrulecolor{black}
Total (per-scene) & ($\sim$ 4-5 min) & - \\

\midrule
\multicolumn{1}{l}{\textbf{\textit{\circlenum{5} Segment-Feature Computation}}} & (per-frame avg.) \\[-0.2ex]
Fine-grained 2D segments & 1.99 s & $ T \propto   n_{\mathrm{frames}}$ \\
Pixel-aligned feature computation & 5.72 s & $ T \propto   n_{\mathrm{frames}}$ \\
Multi-view fusion of segment-features & 0.04 s & $ T \propto S$ \\
\arrayrulecolor{black!10}\midrule\arrayrulecolor{black}
Total (per-scene) & ($\sim$ 10-15 min) & (for 75-150 frames) \\
\midrule
\multicolumn{3}{l}{\textbf{\textit{\circlenum{6} Inference}}} \\[-0.2ex]
New text query / vocab. embedding & 0.61 s & - \\
Search in 3D (similarity computation) & 1.57 ms & - \\
\bottomrule
\end{tabular}
}
\caption{{\textbf{Runtime and Computational Complexity.}}
Reported times are averaged over MultiScan test scenes.
The rightmost column shows whether there is a direct proportionality relationship between the total time per \textit{scene}, vs. other parameters such as the total number of predicted object masks ($M$), total number of predicted part-segments ($S$), and RGB frames in the image sequence $  n_{\mathrm{frames}}$.}
\label{tab:runtime_complexity}
\vspace{-12pt}
\end{table}

\subsection{Discussion and Limitations}
One limitation of our work is the reliance on a simple geometrical over-segmentation method for identifying object \textit{parts}. This is evident from the comparison between Tab.~\ref{tab:multiscan-oracle-joint} and Tab.~\ref{tab:multiscan-joint-part-seg}, where oracle mask experiment yields much higher AP scores than those with predicted part masks, indicating room for improvement in 3D part mask quality.

One might reasonably suggest fusing 2D Semantic-SAM masks instead of obtaining 3D segments directly.
While a few methods such as SAM3D~\cite{sam3d} proposed to perform multi-view fusion of 2D masks from SAM~\cite{segmentanything} to obtain segments in 3D, and presented promising results for \textit{object-level} 3D instance segmentation, our empirical analysis has shown that such methods struggle with fusing inconsistent and small \textit{part-level} masks from multiple views. We repeatedly observed that the multi-view fusion of high-granularity Semantic-SAM~\cite{semanticsam} masks directly in 3D yields noisy segments, and concluded that using a geometrical over-segmentation method is more effective for part segmentation. Nevertheless, there are limitations to the geometrical segmentation method we employ for \textit{part} segmentation, as it relies on surface normals. When multiple part segments share similar surface normals, such as drawers of a wardrobe, this approach struggles to distinguish these scene entities from each other. Additionally, our hierarchical segmentation assumes a reasonably well-reconstructed scene. Highly incomplete point clouds or severe reconstruction noise can degrade segmentation quality, though our semantic merging step helps mitigate over-segmentation issues.

Lastly, our approach is limited to two explicit granularity levels (objects and parts), reflecting the lack of evaluation benchmarks for finer-grained segmentation. We hope future work will address these challenges to enable more granular and flexible hierarchical segmentation.
\section{Conclusion}
\label{sec:conclusion}
We presented a novel open-vocabulary 3D scene understanding approach that extends beyond traditional object-centric queries to enable fine-grained search in 3D environments.
By introducing a hierarchical scene representation, our method segments not only object instances but also object parts and generic attributes.
We validated our approach through extensive experiments and introduced new benchmarks for scene-scale open-vocabulary 3D part segmentation, achieving significant improvements over existing methods. We hope this work inspires future advancements in 3D open-vocabulary segmentation, enabling more flexible handling of scene entities across different levels of granularity.

\ifCLASSOPTIONcaptionsoff
  \newpage
\fi

\bibliographystyle{IEEEtran}
\balance
\bibliography{egbib}
\clearpage

\end{document}